# Machine Learning with Chaotic Strange Attractors

Bahadır Utku Kesgin and Uğur Teğin[*]

Department of Electrical and Electronics Engineering, Koç University, Istanbul, 34450, Turkey

*corresponding author: utegin@ku.edu.tr

**Abstract**

Machine learning studies need colossal power to process massive datasets and train neural networks to reach high accuracies, which have become gradually unsustainable. Limited by the von Neumann bottleneck, current computing architectures and methods fuel this high power consumption. Here, we present an analog computing method that harnesses chaotic nonlinear attractors to perform machine learning tasks with low power consumption. Inspired by neuromorphic computing, our model is a programmable, versatile, and generalized platform for machine learning tasks. Our mode provides exceptional performance in clustering by utilizing chaotic attractors' nonlinear mapping and sensitivity to initial conditions. When deployed as a simple analog device, it only requires milliwatt-scale power levels while being on par with current machine learning techniques. We demonstrate low errors and high accuracies with our model for regression and classification-based learning tasks.

**Introduction**

Current computing methods and hardware limit machine learning studies and applications regarding speed, data resolution and deployed platforms. Particularly, the power consumption of artificial neural networks started to raise questions regarding its impact on the environment. Recent studies indicate that the carbon emissions of training a complex transformer learning model are roughly equivalent to the lifetime carbon emissions of five cars[1], and training a famous language model consumed the energy required to charge 13,000 electric cars fully[2]. Several computing paradigms are proposed for machine learning studies to decrease training times and, therefore, the energy consumption issue. Among them, reservoir computing[3,4] offers a promising path by using nonlinear systems with fixed weights to process information in high dimensional space. Various neuromorphic devices[5] were proposed to surpass chronic performance issues of conventional computing and high-power consumption issues. Optical computing methods[6,7] and electronic memristive devices[8–10] were introduced as powerful reservoir computing platforms. The concept of fixed nonlinear high-dimensional mapping is of usual practice in several areas of machine learning, such as extreme learning machines[11] and support vector machines[12,13].

In machine learning studies, chaotic systems were mainly employed as targets to learn dynamical systems[14–16]. Chaos theory examines deterministic but unpredictable dynamical systems that are extremely sensitive to initial conditions. These systems commonly occur in nature, inspiring art, science, and engineering[17]. Also, chaotic spiking dynamics of neurons have inspired several neuromorphic machine learning applications[18,19]. In the past, chaotic systems were proposed for Boolean computation and data processing, forming the concept of chaos computing. Early chaos computing devices operated one-dimensional chaotic maps to perform logic operations[20,21]. These dynamical systems were also suggested for reservoir computing but used in a stable state just below the bifurcation point, where order transitions to chaos[22]. Operating in a stable state, such systems could not benefit from chaos in learning and information processing for machine learning purposes. Following these attempts, systems with "weakly chaotic" architecture were proposed[23,24]. However, these models and other similar approaches(25) could not demonstrate competent performances[25].

Here, we propose an analog computing method based on controllable chaotic learning operators to perform high-dimensional nonlinear transformations on input data for machine learning purposes. Our method benefits circuits designed to compute chaotic strange attractors for reservoir computing purposes, as demonstrated in Fig.1. Since minor differences amplify and evolve, chaotic transformation processes information and improves performance for machine learning tasks. While previously reported physical reservoir computing hardware lacks flexibility, we introduce a controllable model by increasing overall versatility. Achieving this versatile platform allows us to enhance overall learning accuracy for various learning tasks through optimization. Our computing method intrinsically offers smaller footprints with power consumption levels as low as a milliwatt scale while preserving high accuracies. By providing complex and chaotic dynamics for the nonlinear transformation of data, our model performs on par with neural networks operating on conventional computing platforms. We present the generalizability of our approach by testing a wide variety of machine learning tasks, including image classification, and achieve high accuracies, reaching up to %99 for several tasks. Later, we explore how sensitivity to initial conditions in chaotic attractors improves learning accuracy and determines the power consumption required for training. Our method is a controllable, chaotic analog learning device that offers versatile and sustainable machine learning without compromising learning performance.

**Results**

*Input / Output encoding and selection of the optimal attractor*

As chaotic systems are extremely sensitive to initial conditions, we anticipate that the input method is highly correlated with output accuracy. We decide to input our data as initial conditions of the attractor. As we scale our data using z-score normalization, the initial conditions we use as inputs land in a scale that does not vitiate the physical model (see Supplementary Material for details). After the chaotic transformation is applied to our samples, we feed the transformed matrix to the regression or classification algorithm.

The pattern and average divergence pace between each chaotic attractor's close points are distinctive properties. We select six different chaotic attractors to evaluate how these unique properties translate into machine learning. We employ a nonlinear regression task on a randomly generated Sinus cardinal dataset (see Methods). We select the well-known Lorenz attractor[26], Rössler attractor[27], Burke-Shaw system[28], Sprott attractor[29], Chen's system[30], and Chua's Circuit[31] for this test. Our test attractors transformed randomly generated points for one hundred iterations and tried to predict

Sinus Cardinal function values corresponding to the transformed sample. After recording the lowest root mean squared error (RMSE) amongst iterations, we sort each result from smallest to largest RMSE value. Lorenz attractor was the most successful attractor with a RMSE of 0.143. We decide to proceed with further tests only using the Lorenz attractor and using the iteration with the lowest error after 100 iterations (see Supplementary Material for details).

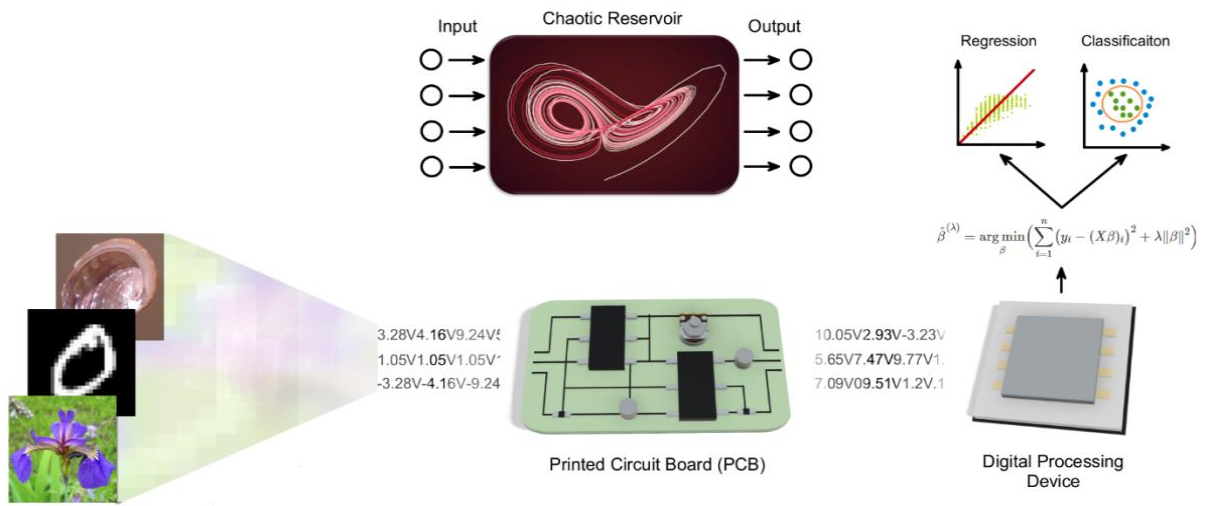

**Figure 1:** Schematic displaying the architecture of our model. Input values are encoded as initial voltages for the circuit performing analog computation of Lorenz attractor. After the chaotic transformation of data, output voltages are transferred to a processing device as reservoir output. Via the device, the last layer is performed, mainly ridge regression and classification, completing the learning process.

*Sinus Cardinal regression*

To assess the potential performance of machine learning with chaotic attractors, we run a simple regression task on a dataset of randomly generated samples and their values after the Sinus Cardinal function. In aforesaid benchmarking tests, we measure the vanilla RMSE of Lorenz attractor as 0.143. We apply the Bayesian Optimization algorithm to determine the best values for Lorenz system parameters to minimize error and improve model performance. After completing three separate optimizations, we select the values that lead to minimum error ($\sigma = 10$, $\beta = 8/3$, and $\rho = 97$). We use these coefficients in further tasks except the Abalone dataset, where we applied a separate optimization. After the optimization, an RMSE of 0.105 is achieved.

To further increase our model performance, we add another layer that will apply the chaotic transformation to the input variable. First, two parallel Lorenz Attractor layers with different $\rho$ values transform the same input simultaneously. These two distinct outputs are concatenated into a single matrix, and this matrix undergoes the learning process. Using two attractors, we increase the dimensionality and benefit from different $\rho$ values (see Largest Lyapunov exponent and accuracy). Keeping $\sigma$ and $\rho$ as constants, we apply Bayesian Optimization to determine optimal ρ values for our transformers. After the optimization process, we decrease model RMSE down to 0.03. (see Supplementary Material for the figure)

*Abalone dataset*

Moving on with a relatively more complex and multivariable regression task, we test our chaotic model in the abalone dataset. This dataset, taken from ref[32], is composed of the eight physical measurements of sea snails and their ages. We normalize the ages on a scale between 0 and 1. We apply z-score normalization and deploy chaotic transformation with a single transformer to every single variable. We use Bayesian optimization to find the optimal parameters of the Lorenz transformer. After optimization, we achieve remarkable accuracy (RMSE 0.072884) with parameters: $σ = 10$, $β = 2.667$, $ρ = 64.917$. (see Supplementary Material for the figure)

*Iris dataset*

We move on with classification tasks to challenge our model. The Iris dataset is one of the classical datasets that assess linear and nonlinear classification abilities. The dataset from ref[33] consists of four physical measures of iris flowers from three distinct species. While one class, iris-setosa, is linearly separable from the other two classes, iris-versicolor and iris-virginica require nonlinear applications to be separated. We employ Ridge classification as the last layer because it is a simple and linear method that is fast to execute. Changing the usual method for visualizing classifier decision boundary, we use Linear Discriminant Analysis (LDA) to raw and transformed data (see figure 2). Using LDA, we retrieve 2D matrices for raw and transformed data and perform Ridge classification to these 2D matrices. A high accuracy of 97,78% is achieved, gaining about 18% over model accuracy before chaotic transformation (80,00%). After chaotic transformation, samples that belong to linearly non-separable classes (iris-versicolor and iris-virginica) all clustered almost perfectly (see Figure 2). As a result, the linear classifier we utilize can make an almost perfect classification. We also test other classifiers for benchmarking (see Methods and Supplementary Material Table 2 for details). A drastic increase in test accuracy of linearly inseparable classes is demonstrated in confusion matrices (see Figure 3).

*Liver disorders dataset*

For this classification task, we test our methods in the liver disorders dataset. This dataset, taken from ref[34], comprises 12 features in blood samples taken from healthy people, hepatitis patients, fibrosis patients, or cirρsis patients. After obtaining an even dataset (see Methods for details), we employ the same chaotic transformation method to our features. With chaotic transformation, we report an increase in the ridge classifier accuracy by about 11% from 81.71% to 92.82% and achieve an accuracy of 98.84% with Linear SVM (see Supplementary Material). With chaotic transformation, like previous results, classes are well-clustered and decision boundary lines are easier to draw (see Figure 2). Also, substantial improvement in the accuracies of every single class is displayed in confusion matrices (see Figure 3).

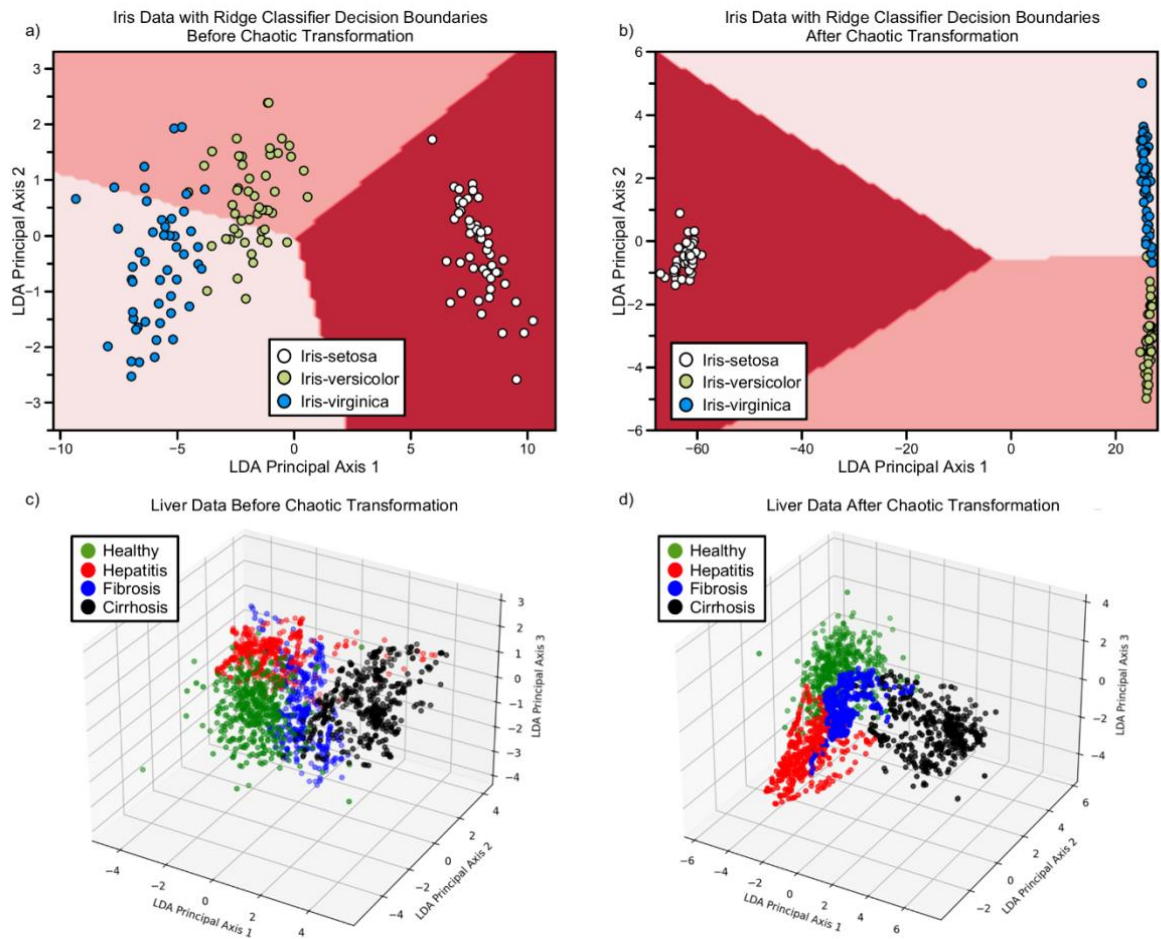

**Figure 2:** Impact of chaotic nonlinear transformation on the decision boundaries and the data points. **a,b,** Decision boundary of ridge classifier in Iris dataset before (**a**) and after (**b**) the chaotic transformation. **c,d,** Distribution of datapoints of Liver dataset before (**c**) and after (**d**) the chaotic transformation.

*MNIST dataset*

We test our model for image classification after proving strong performance in numerical datasets. MNIST dataset[35] contains 70,000 samples (60,000 training, 10,000 testing) of 10 handwritten digit classes. For this task, 28x28 images are flattened without any normalization, and a fast algorithm for dimensionality reduction (see Methods for details) is employed as a form of preprocessing. After reducing dimensions of each flattened images from 1x784 to 1x7, we perform classification and set a baseline accuracy. After chaotic transformation, the accuracy of this Ridge classifier increase 81.42% to 95.42%. Such a significant increase in accuracy highlights the effect of chaotic nonlinear transformation one more time.

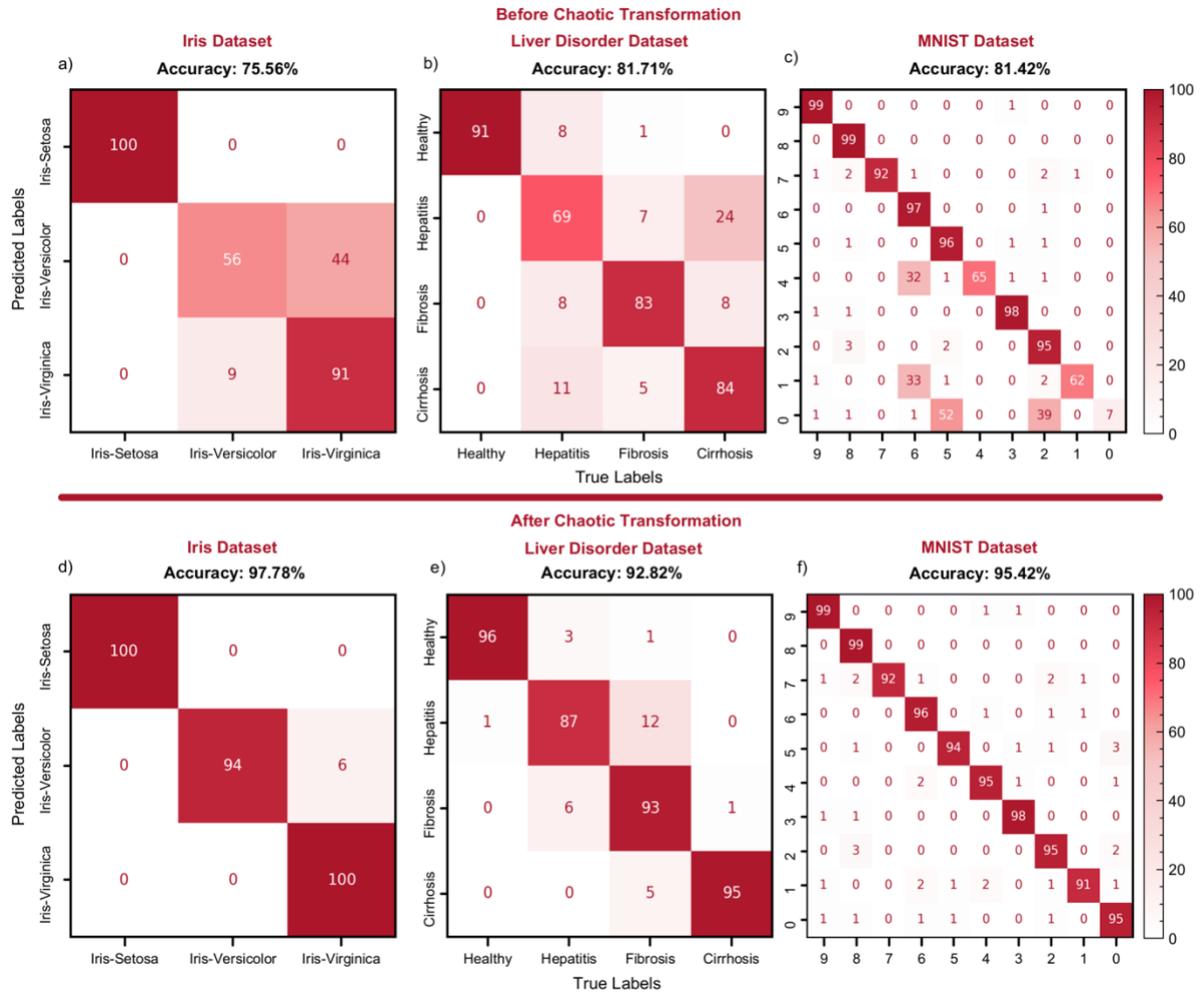

**Figure 3:** Confusion matrices of Ridge classifier accuracy in three classification tasks before and after chaotic transformation. The upper row (**a, b, c**) represents accuracies before chaotic transformation was applied and the lower row (**d, e, f**) represents accuracies after chaotic transformation. Confusion matrices of each dataset are represented as follows: Iris dataset (**a, d**), Liver Disorders Dataset (**b, e**), and MNIST Dataset (**c, f**). Confusion matrices are normalized row-wise (see Methods for details).

*Largest Lyapunov exponent and learning accuracy*

Next, we investigate the impact of sensitivity to initial conditions on our model's performance in machine learning tasks. We set the Largest Lyapunov Exponent ($\lambda$)(LLE)[36] to measure the pace of divergence in a chaotic system. An LLE that is larger than 0 indicates a chaotic system, and a larger LLE corresponds to faster diverging points.

In this test using the Liver Disorder dataset, we study a chaotic transformation with $\rho$ values ranging between 1 to 100. Then, we record the best accuracy of Linear SVM. We evaluate the LLE of Lorenz attractor with $\rho$ value in the range from 1 to 100. When compared with a non-chaotic ($\rho$ = 2) and chaotic but less divergent model ($\rho$ = 28), the optimized model (ρ = 97) demonstrates higher accuracies in every single class (see Figure 4).

We also demonstrate a positive statistical relationship between the Largest Lyapunov Exponent and model accuracy after running Welch's t-test and Pearson's R-value test (see Supplementary Material for details). It should be noted that as these dynamical systems evolve, while we benefit from the divergence, as mentioned earlier at early iterations, the attractor transforms the data, becoming unlearnable after a particular stage.

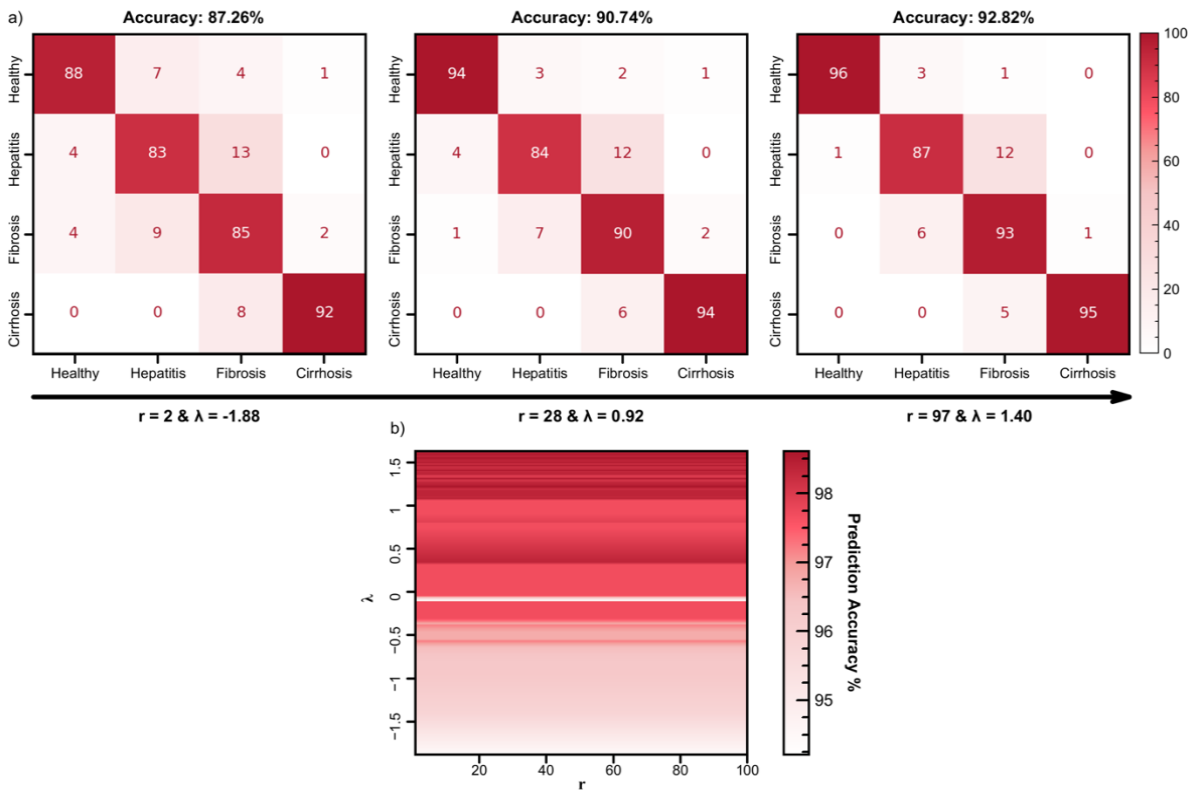

**Figure 4:** Relation between sensitivity to initial conditions and model accuracy. **a,** Confusion Matrices of Ridge classifier in Liver Disorder Dataset on three states of Lorenz transformer: non-chaotic (stable), chaotic, and more chaotic **b,** Color map visualization of P value (x-axis) LLE (y-axis) and accuracy of Linear SVM in liver disorder dataset (color values).

*Circuit simulations*

Encouraged by our model's impressive performances, we study its analog implementations with circuit simulations using a specific circuit designed for the analog computation of the Lorenz attractor[37]. After running the circuit and performing chaotic transformation to the data, we use a decision layer like our previous tests. We tune the analog computing achieved via the circuit by changing the resistance of a resistor and adjusting the ρ value. Alternatively, a digital potentiometer can be utilized to actively set the effect of chaotic data transformation in the circuit.

Our circuit simulations delivered the same performances with numerical test results when ρ is set to 97, thus proving the feasibility of our proposed analog learning model (See Supplementary Material Table 1). In circuit simulations, we calculate the total power consumption of our analog chaotic systems. A single analog unit consumes about 350 milliwatts (see Supplementary Material for details) to perform the chaotic transformation to data.

**Discussion**

The findings of this study present a promising computing platform for the field of machine learning. The study introduces a novel method that has demonstrated effectiveness in various machine learning applications. It significantly improves power consumption for image and numerical classification tasks, using a straightforward linear last layer following a chaotic nonlinear transformation. This methodology, showcased in the context of MNIST, Liver Disorder, Iris, Abalone, and Sinus Cardinal datasets, not only enhances accuracy but also maintains input data integrity and permits flexible adjustments of model parameters and architecture.

One intriguing aspect of this study is the integration of circuit simulations to validate the practicality of the analog chaotic reservoir computing paradigm. This approach also enables an in-depth examination of the relationship between the Largest Lyapunov exponent of the chaotic transformer and overall model accuracy. Moreover, the circuit architecture's speed and power efficiency on a milliwatt scale hold promise, particularly in light of contemporary concerns regarding energy consumption in machine learning applications.

The Lorenz attractor, serving as the primary chaotic transformer in this study, emerges as a noteworthy element, showcasing remarkable performance in clustering and pattern recognition. The potential for further research in related areas, particularly in image segmentation using chaotic pattern recognition, is a direction that warrants exploration. The study also highlights how optimizing the chaos parameter, ρ, can lead to modest yet appreciable increases in model test accuracy. The positive correlation between model accuracy and the Largest Lyapunov Exponent raises intriguing possibilities for future research. Our method opens the door to various opportunities for further investigation, particularly in the realm of neuromorphic architectures that can harness chaos as a computational element. Similar chaotic computing techniques can be realized with silicon-on-insulator technology for chip-size footprints. Such architectures may offer innovative solutions and insights for advancing the field of machine learning.

## Methods

In our method, we compute the following set of ordinary differential equations for the Lorenz attractor to transform our data[26]:

$$\frac{dx}{d\tau} = -\sigma x + \sigma y$$

$$\frac{dy}{d\tau} = -xz + \rho x - y$$

$$\frac{dz}{d\tau} = xy - \beta z$$

where we use coordinates x, y, and z for both input and output recording, and use parameter ρ to adjust chaos. We created a Python code using the NumPy[38] library that will iterate the ordinary differential equations of chaotic strange attractors in time using the Runge-Kutta method[39,40]. Due to high dimensionality, each variable is given to the simulation code as an (x, y, z) vector in (variable, 1.05, -variable) format. This code is then used to perform reservoir computation on the given input. Due to the high dimensionality of strange chaotic attractors, every one-dimensional predictor is transformed into a three-dimensional vector. Besides the exception of the Iris dataset, all the output vectors are used for the learning process. In the Iris dataset, after the transformation of all samples is complete, the Linear Discriminant Analysis method is applied to data before the final layer to demonstrate the decision layers and the learning process is consistent. A timestep of $10^{-2}$ is used to simulate strange chaotic attractors. Unless stated otherwise, the coefficients of used attractors were in their author-suggested values for the attractor benchmark test.

Excluding Sinus Cardinal data, every other dataset used was normalized using z-score normalization with a standard deviation of samples equal to one before being transformed with chaotic attractors. The Sinus Cardinal dataset is synthetically created and not normalized, with the predictor being randomly generated 2048 samples in the range of [-pi, +pi] and target values being the Sinus Cardinal function of generated samples. The Liver Dataset is an uneven dataset, which may result in imbalanced learning, and to prevent this, we used the Python implementation of the Synthetic Minority Oversampling Technique to upsample the Liver dataset evenly. In the MNIST dataset, we flattened the dataset and applied dimensionality reduction using Uniform Manifold Approximation and Projection for Dimension Reduction[41](UMAP).UMAP reduced the predictor size to 1/112 of the original data (784 to 7). Dimensionality reduction lasted about two minutes. A ratio of 80% training set to 20% test set was used to divide the datasets into training and test sets. Only for the Iris dataset, a ratio of 70% training set and 30% test set was used to divide the dataset. In all displayed results, datasets are set to training and test tests using random state zero.

For the regression tasks (Abalone and Sinus Cardinal), predictors of every sample are transformed with our code, and a simple Linear Regression algorithm is implemented as the final layer that completes the learning process. For classification tasks (MNIST, Liver Disease, and Iris), following the same transformation process, the Ridge Classification, Linear Kernel Support Vector Machine (SVM), Polynomial Kernel SVM, Gaussian Kernel SVM, K-Nearest Neighbors, and Multilayer Perceptron Classifier algorithms are used as the last layer. All the last layers are implemented using the Scikit-learn[42] package. Unless stated otherwise in the results or methods section, all classifiers are utilized using their default method in the scikit-learn package. The multilayer perceptron classifier utilized in the study comprises a learning rate of $10^{-3}$, a tangent hyperbolic (tanh) activation function, and three hundred hidden neurons. Confusion matrices are normalized over true predictions (row-wise), and decimal numbers are rounded to the nearest whole. The table results show the standard deviation of accuracies after 20 separate dataset splits.

For the Chaos and Learning test, we estimated LLEs using the method proposed by Rosenstein et al.[43]. We utilized the MATLAB built-in function for to estimation Lyapunov Exponents. We measured local LLE between iterations 1 and 200 as these iterations were our range. We decided to make parameters $\sigma$ and $\beta$ as fixed variates ($\sigma = 10$, $\beta = 8/3$). We decided to keep these parameters unchanged due to the high sensitivity to the initial conditions of the Lorenz Attractor, which would complicate testing. We employed the Linear SVM with MATLAB implementation for chaos and learning tests. We utilized the SciPy[44] library functions of the given statistical significance tests.

For the circuit simulations, we modified the schematic of the circuit that performs the analog computation of the Lorenz system[37] to be able to input initial conditions. We then converted this schematic to a netlist file that we will feed to LTSpice (see Supplementary Material for Circuit Schematic). This netlist file consists of the circuit structure and the commands to regulate the circuit simulations. Identical to the numerical simulations, we set the timestep of the circuit simulation to 10μs and iterated the circuit one thousand times. Afterward, we created a Python code to work simultaneously with the LTSpice simulation engine and perform parallel circuit simulations. For every variable in a sample in the dataset, this code initiates a circuit simulation after modifying the initial conditions as the variable's value. Results of the simulations are stored in a ". raw" format file that will require another Python code to extract output values. This code we created retrieves one thousand iterations of every sample out of the result files and creates a matrix of output values. To complete the learning process, values are sliced iteration-by-iteration from the matrix, and the same final layers in the numerical simulations are applied to the sliced values. We retrieved power consumption data by slicing power dissipation data from the same result files.

**Data and code availability**

Datasets that contain raw information are available in references[32–35]. All numerical and circuit simulation test data and code are available upon reasonable request.

**Author contributions**

B.K. performed simulations and tests, U.T supervised and directed the project. All the authors participated in the data analysis and the manuscript's writing process.

**Competing interests**

The authors declare no competing interests.

# Supplementary Material of

# Machine Learning with Chaotic Strange Attractors

Bahadır Utku Kesgin and Uğur Teğin[*]

Department of Electrical and Electronics Engineering, Koç University, Istanbul, 34450, Turkey

*corresponding author: utegin@ku.edu.tr

*Creation of Input / Output Encoding*

Chaotic attractors are multidimensional systems however the data we process is one-dimensional. Therefore, the creation of a stable encoding method is crucial for model stability. Initially we insert the data as the initial condition of only one dimension and kept the other two dimensions as constant variables for every predictor. We test this pipeline with Lorenz attractor, which is three-dimensional with initial conditions of (variable, 1,1). While left as a dummy variable with a value of 1, the y and z coordinates are also processed and produce unique outputs every iteration. Therefore, we start to use the output value of the dummy variables in the learning phase to benefit from the increased dimensionality and unique values of the dummy variable. In the end, our chaotic transformation method expands dimensionality from 1D to 3D for every input. After chaotic transformation with Lorenz attractor is applied to our randomly generated samples, we feed the output matrix to the regression algorithm. This process is repeated every iteration for one hundred iterations, and the best accuracy/lowest error is recorded. We record the minimum Root Mean Squared Error (RMSE) of the input method (variable, 1, 1) as 0.11. We perform series of tests with different combinations and achieved the lowest error among other tested combinations using the initial conditions (variable, 1.05, -variable) with an RMSE of 0.10.

| Attractor | Sinus Cardinal Regression Minimum Root Mean Squared Error (RMSE) | Parameters |
|---|---|---|
| Before Transformation | 0,346368 | N/A |
| Lorenz Attractor | 0,149351 | a = 10  b = 2.667  c = 28 |
| Chua's Circuit | 0,187065 | a = 9  b = 100/7  c = 8/7  d = 5/7 |
| Chen's System | 0,201063 | a = 60  b = 2.667  c = 97 |
| Burke-Shaw Attractor | 0,234342 | a = 10  b = 4.272 |
| Sprott Attractor | 0.240657 | a = 2.07 b = 1.79 |
| Rössler Attractor | 0,301912 | a = 0.2  b = 0.2  c = 5.7 |

**Supplementary Table 1:** Linear Regression results Sinus Cardinal regression for six selected chaotic strange attractors and before transformation

## Regression and classification results

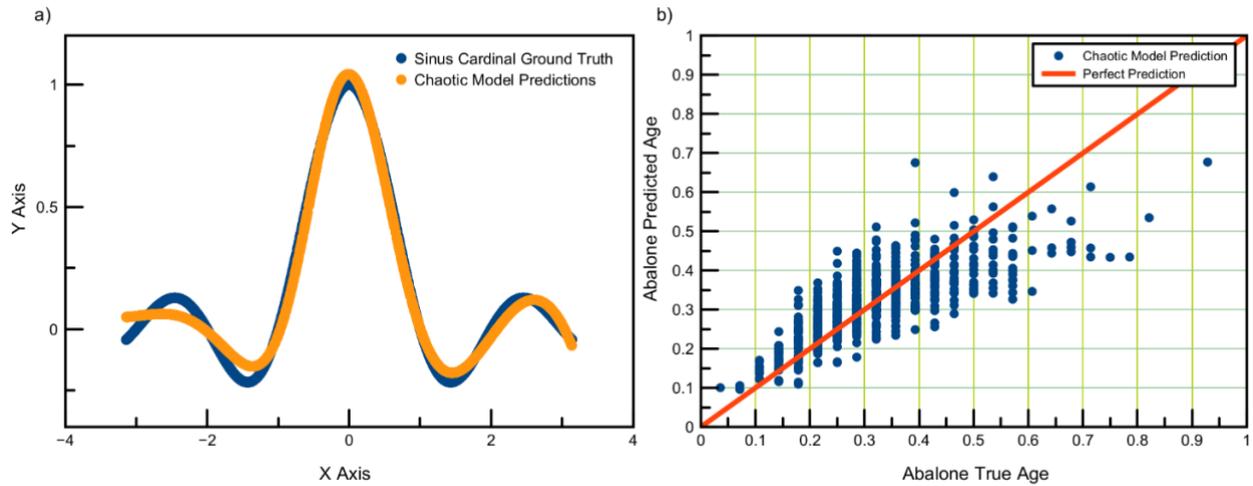

**Supplementary Figure 1:** Linear Regression results after the application of chaotic transformation with Lorenz attractor. a) Measured linear regression results after the application of two parallel chaotic transformations with ground truth b) Measured linear regression results of single chaotic transformation to abalone dataset.

| Classifier | Liver Dataset Test Accuracies | | | Iris Dataset Test Accuracies | | | MNIST Dataset Test Accuracies (After UMAP) | | |
|---|---|---|---|---|---|---|---|---|---|
| | Before Transformation | Numerical Simulations (Iteration: 32) | Circuit Simulations (Iteration: 290) | Before Transformation | Numerical Simulations (Iteration: 98) | Circuit Simulations (Iteration: 251) | Before Transformation | Numerical Simulations (Iteration: 7) | Circuit Simulations (Iteration: 207) |
| Ridge Classifier | 81,71% | 92,82% (0,85) | 91,66% (1,42) | 80,00% | 97,78% (2,16) | 97,78% (1,78) | 81,42% | 95,42% | 95,42% |
| Linear SVM | 91,20% | 98,84% (0,31) | 96,52% (0,77) | 97,78% | 97,78% (3,02) | 95,55% (2,95) | 95,56% | 95,61% | 95,59% |
| Polynomial SVM | 90,74% | 88,42% (1,05) | 93,98% (1,03) | 95,56% | 77,78% (8,74) | 93,33% (2,96) | 95,62% | 95,58% | 95,60% |
| Gaussian SVM | 88,43% | 83,79% (1,53) | 92,59% (1,16) | 97,78% | 60,00% (7,47) | 97,78% (2,21) | 95,57% | 95,53% | 95,56% |
| K-Nearest Neighbors | 97,69% | 95,60% (0,79) | 97,91% (0,53) | 95,56% | 91,11% (4,21) | 97,78% (2,17) | 95,64% | 95,65% | 95,65% |
| Neural Network | 93,39% | 97,22% (0,81) | 99,30% (0,58) | 97,78% | 93,33% (3,11) | 97,78% (2,21) | 95,66% | 95,62% | 95,65% |

**Supplementary Table 2:** Results of test accuracies of different classification algorithms in all three datasets before chaotic transformation, after numerical simulations of chaotic transformation and after circuit simulations of chaotic transformation. Chaotically transformed data is split into training and test sets for twenty times, standard deviations for every classifier are given in the parentheses. Accuracies after transformation are given in best iterations of the model.

### Computation of largest Lyapunov exponents

We use MATLAB's built-in function to calculate largest Lyapunov exponent. This function estimates the Largest Lyapunov Exponent using algorithm proposed by Rosenstein et.al[1]. We iterate the attractor using our iterator and apply Phase Space Reconstruction[2,3] to output data. Estimator function is applied to one-dimensional reconstructed data and estimated Lyapunov Exponents are recorded. We measure local Lyapunov exponent between our operating range. We use MATLAB for only this section due to MATLAB code being drastically faster.

### Statistical significance tests

We gather all data retrieved in the MATLAB code to a CSV file to interpret and feed the file to a Python code that will apply Welch's t-test and Pearson's R-value test to Lyapunov Exponents and Model Accuracies. Pearson test indicated a positive correlation with a r-value of 0.84 between these two values. In Welch's t-test, we obtain a t-value of 1.35 and our null hypothesis passes with a significance level of 0.1 (80% confidence), but the fails in significance level of 0.05 (90% confidence), pointing to a mild correlation. We use Welch's t-test over Student's t-test because the means of the two data are drastically different. The degree of freedom in the t-test was two hundred.

### Circuit simulations

We modify the circuit schematic from ref[4] in order to be able to give direct inputs. We add two voltage sources to two independent nodes (y and z) with switches to give initial input to the circuit. All constants are scaled to resistor values in 1MΩ, for example, ρ equals 1 MΩ /(R8+R9) and beta equals 1 MΩ /(R4+R5). Changing the value of R9 we modified the ρ value in our experiments. For example, when the resistance of R9 is equal to 33 kΩ, the circuit computes the Lorenz attractor with a ρ value of 28 and when the resistance of R9 is equal to 7.601kΩ, the circuit computes the Lorenz attractor with a ρ value of 97.

In the PCB version multiplications are conducted with AD633 analog multiplier and TL074 op-amps that draw +/- 15V are utilized for simple operations. Our model consists of 2 AD633 analog multipliers and three LT074 operational amplifiers. In LTSpice simulations we didn't include these components for convenience but gave commands that are identical to the operations performed by AD633 and LT074.

When calculated in maximum supply current and voltages, a single AD633 consumes 108 mW and a single LT074 consumes 45 mW. Considering the amounts of components, we use; a single chaotic transformer consumes 351mW at maximum. It should also be noted that this transformer will not always run at maximum voltage due to the oscillatory behavior of chaotic systems, therefore total power consumption would be less than this projection.

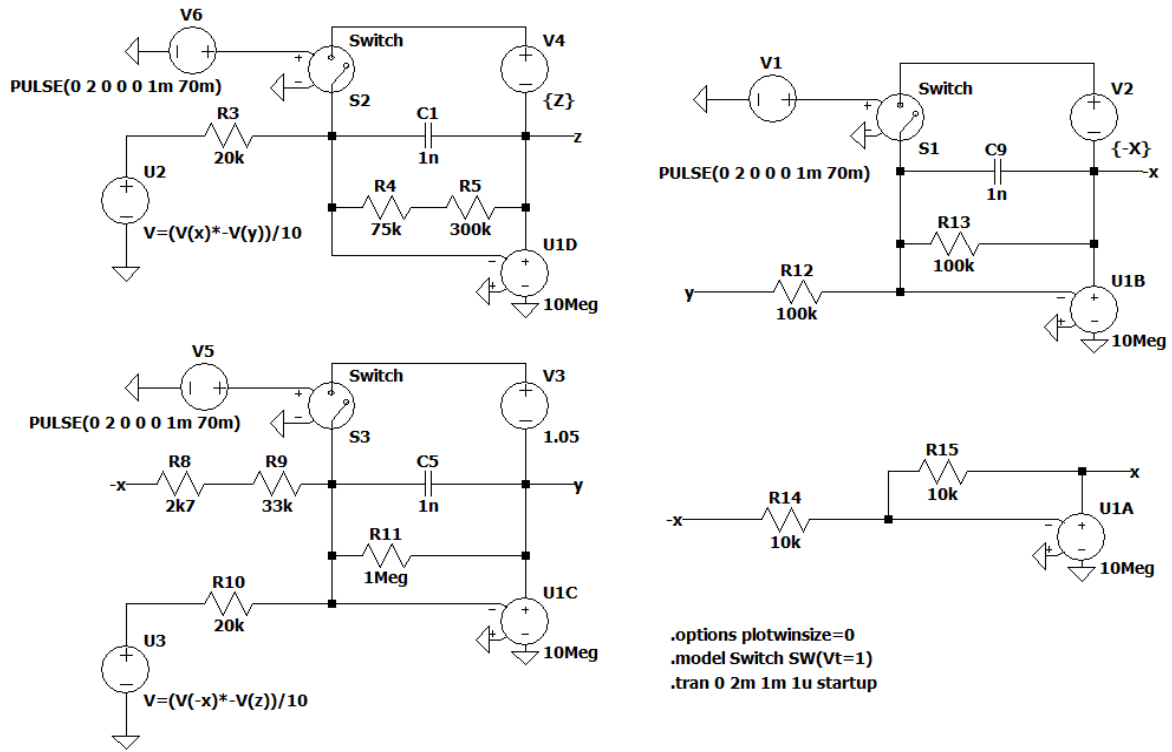

**Supplementary Figure 2:** Circuit schematic used in the LTSpice simulations with along simulation commands.

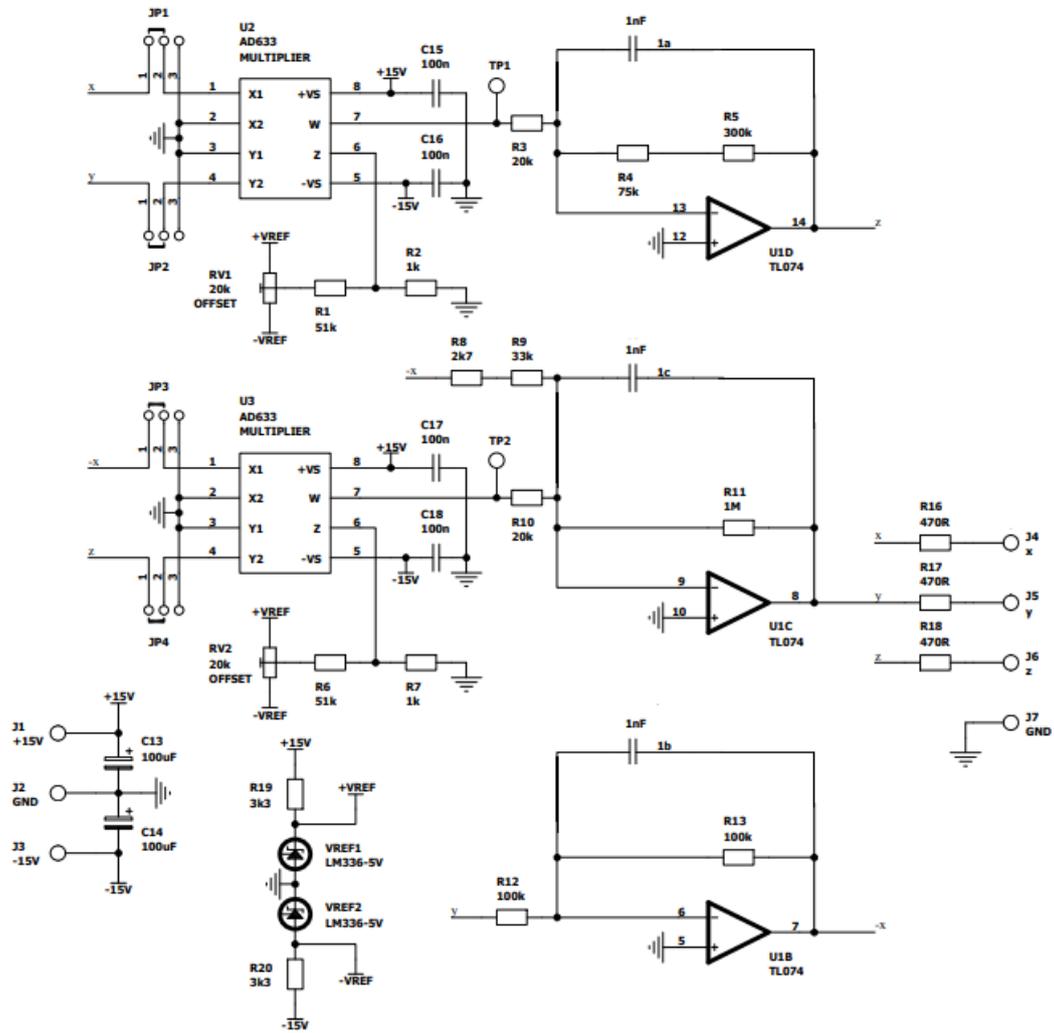

**Supplementary Figure 3:** Initial circuit schematic designed to use for actual analog implementation of Lorenz attractor.

**Supplementary Material References**